%% file: iclr2022_conference.tex
\documentclass{article} 
\usepackage{iclr2022_conference,times}

\usepackage{hyperref}
\usepackage{url}

\input{math_commands.tex}

\usepackage{algpseudocode}
\usepackage{algorithm}

\usepackage{graphicx}
\graphicspath{ {./images/} }
\iclrfinalcopy

\title{Using Multiple Self-Supervised Tasks Improves Model Robustness}

\author{Matthew Lawhon, Chengzhi Mao \& Junfeng Yang \\
Department of Computer Science\\
Columbia University\\
New York, NY 10027, USA \\
\texttt{matthew.lawhon@columbia.edu, \{mcz, junfeng\}@cs.columbia.edu}
}

\begin{document}

\maketitle

\begin{abstract}
Deep networks achieve state-of-the-art performance on computer vision tasks, yet they fail under adversarial attacks that are imperceptible to humans. In this paper, we propose a novel defense that can dynamically adapt the input using the intrinsic structure from multiple self-supervised tasks. By simultaneously using many self-supervised tasks, our defense avoids over-fitting the adapted image to one specific self-supervised task and restores more intrinsic structure in the image compared to a single self-supervised task approach. Our approach further improves robustness and clean accuracy significantly compared to the state-of-the-art single task self-supervised defense. Our work is the first to connect multiple self-supervised tasks to robustness, and suggests that we can achieve better robustness with more intrinsic signal from visual data.
\end{abstract}

\section{Introduction}



Deep learning architectures achieve state-of-the-art and often superhuman performance across a wide variety of vision tasks \citep{croce2020reliable}. Despite this, as first noticed in 2014, they remain vulnerable to \textit{adversarial attacks}: visually imperceptible perturbations that cause easily classifiable images to be misclassified with high confidence by otherwise state-of-the-art machine learning algorithms \citep{szegedy2014intriguing}. This results in unpredictable behavior in edge-cases, contrived examples and examples unrepresented in training data. The inability to address this sufficiently is a leading hurdle to deploying deep learning solutions to human safety and well-being critical applications like autonomous transportation and health-care.

Though there is a large line of research into \textit{adversarial training}, how we can train networks to resist adversarial attacks, training time defenses can be very computationally expensive and it is difficult to provide guarantees for all possible attack methods \citep{tramr2020ensemble}. Empirically, \textit{unrestricted white box attacks}, in which an adversary has complete, unrestricted access to the network it is trying to corrupt, have been found very difficult to resist via adversarial training. \citet{mao2021adversarial} propose to adapt to adversarial attacks at test time by restoring the performance of a selected self-supervised task, however, the adaption method significantly reduces the clean accuracy after adaptation because it over-fits to a single self-supervised task.

\begin{figure}
    \centering
    \includegraphics[scale = 0.4]{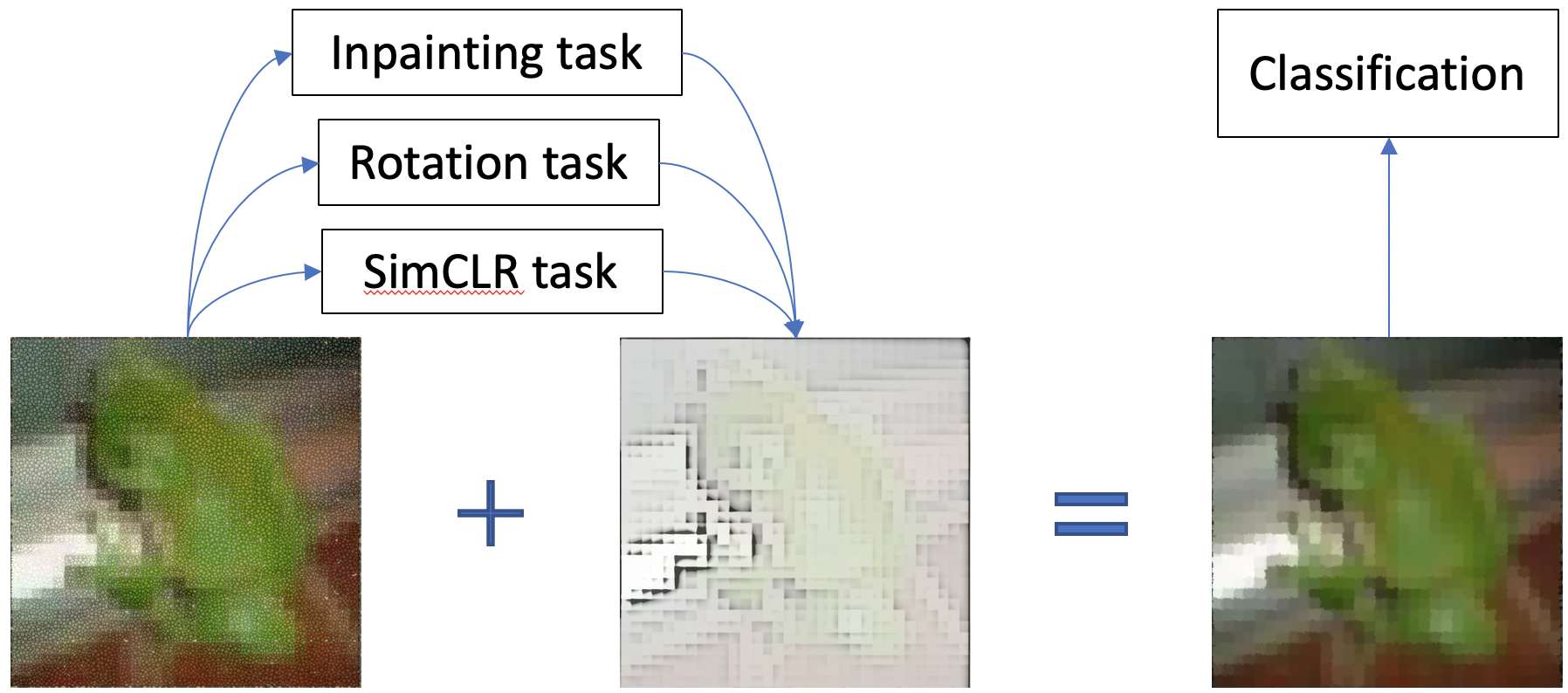}
    \caption{Given a possibly attacked image, in our approach we find some small reverse perturbation to minimize multiple self-supervised learning task losses and recover structure in the image. Adding this reverse perturbation to our image, we are able to improve classification accuracy on attacked images with minimal disruption to classification accuracy on unattacked images.}
    \label{fig:my_label2}
\end{figure}

In this paper, we propose to mitigate the over-fitting problem present in \citet{mao2021adversarial}'s reversal method by using multi-task learning. Our key insight is that different self-supervised tasks, none of which require labels, capture different aspects of the intrinsic structure of images. By restoring the performance of attacked images on multiple self-supervised tasks, we can recover a larger set of features that have been corrupted by an adversarial attack. In addition, \citet{mao2020multitask} show that it is harder to simultaneously attack multiple tasks, suggesting that this methodology is also resistant to an adaptive attacker who has full access to our reversal methodology and attempts to optimize their attack knowing our reversal methodology in place.

Using three self-supervised tasks, our approach yields statistically significant improvement upon the state-of-the-art defense. On the CIFAR-10 dataset using \citet{carmon2019unlabeled}'s robustly trained baseline model, we achieved a 1.1\% improvement in classification accuracy of unattacked images, and a 0.3\% improvement in classification accuracy of PGD attacked images compared with the state-of-the-art results from \citet{mao2021adversarial}. Our work suggests that deep computer vision models' robustness can be improved by ensuring that they leverage the rich intrinsic structure of image data.

\section{Related Work}

\subsection{Adversarial Attacks}
Early work in this field demonstrated the susceptibility of neural networks trained to solve computer vision tasks to human-imperceptible amounts of noise that would result in high-confidence misclassifications \citep{szegedy2014intriguing}. Notable early adversarial attack methods include the Fast Gradient Sign Method (FGSM) from \citet{goodfellow2015explaining}, and Projected Gradient Descent (PGD) from \citet{madry2018towards}, in which the attacker attempts to maximize the loss function of a classification network within the local $\epsilon$-neighborhood of a particular example. In this paper, we assume white box access to the network and reversal procedures, and attempt to optimize attacker performance using the same minimax optimization setup presented initially by \citet{mao2021adversarial}.

\subsection{Self-supervised Learning}
Image data contains rich intrinsic structure that can be used in learning image representations. In Self-supervised learning for images we use deep architectures to learn unsupervised tasks including canonical examples like rotation prediction, inpainting and contrastive predictive coding \citep{gidaris2018unsupervised, ,pathak2016context, chen2020simple}. \citet{mao2021adversarial}'s first noted the transferability of adversarial attacks on classification, a supervised task, to contrastive learning. While we can't repair an attacked image by minimizing the classification loss since we don't have a ground truth label, we can still minimize loss for self-supervised tasks. They show the effectiveness of reversing adversarial attacks by minimizing contrastive loss thereby implicitly strengthening the inherent structure in the image. 

\subsection{Multitask Learning}
In multitask learning, we attempt to learn multiple related tasks at once using a shared architecture. Heuristically, this leverages the fact that much of the representational information needed to solve related tasks is shared \citep{caruana_multitask_1997}. \citet{mao2020multitask} provide theoretical and empirical results concerning multitask learning's ability to enhance robustness of single-task and multi-task attacks. In this work, we incorporate these theoretical and empirical observations by leveraging multiple self-supervised tasks to repair potentially attacked images, instead of one. 

\section{Our Approach}

In our approach we improve \citet{mao2021adversarial}'s results by migrating their self-supervised learning based image repair to a multi-task learning approach using multiple self-supervised tasks. In this paper we experimented with SimCLR, Inpainting and Rotation prediction tasks, but note that this is a somewhat arbitrary and preliminary choice that would benefit from additional investigation in future work \citep{gidaris2018unsupervised, pathak2016context, chen2020simple}.

\subsection{Attack Model}

We use a standard attack model in which for a given image $\vx$, classifier $F$ and its loss function $\mathcal{L}_c$ (we use cross-entropy loss, defined as $\mathcal{L}_c(\vx, y) = H(F(\vx), y)$), norm parameter $p$ ($\infty$ here) and small $\epsilon$, the attacker searches for an adversarial perturbation $\vx_a$ where $$\vx_a = \arg\max_{\vx_a} \mathcal{L}_c(\vx + \vx_a, y): \|\vx_a\|_p<\epsilon$$

\subsection{Reverse Model}

\begin{algorithm}[t]
\caption{Multi-task Learning Reverse Attack}
\label{algorithm: SSLattack}
\begin{algorithmic}[1]
\Require Potentially attacked image $x$, step size $\eta$, number of iterations $K$, a classifier $F$, reverse attack bound $\epsilon_v$, contrastive loss $\mathcal{L}_s$, roation loss $\mathcal{L}_r$ and inpainting loss $\mathcal{L}_i$.
\Ensure Class prediction $\hat{y}$
\State $x' \gets x+n$, where $n$ is the initial random noise
\For{$k=1,...,K$}
    \State $L = \mathcal{L}_s(\vx)+\mathcal{L}_r(\vx)+\mathcal{L}_i(\vx)$
    \State $\vx' \gets \vx' - \eta ( \mathcal{L}_s(\vx)/L)\nabla_{\vx} \mathcal{L}_s(\vx)$
    \If{$k \mod 2 = 0$}
        \State $\vx' \gets \vx' - \eta ( \mathcal{L}_r(\vx)/L) \nabla_{\vx}\mathcal{L}_r(\vx)$
    \Else
        \State $\vx' \gets \vx' - \eta( \mathcal{L}_i(\vx)/L) \nabla_{\vx}\mathcal{L}_i(\vx)$
    \EndIf
    \State $\vx' \gets \Pi_{(\vx,\epsilon_v)} \vx' $  which projects the image back into the bounded region.
    \State $\vx \gets \vx'$
\EndFor
\State Predict the final output by $\hat{y}=F(\vx')$
\end{algorithmic}
\end{algorithm}

We observe that often $\vx_a$ is designed in such a way as to disrupt the inherent structure of the resultant image $\vx+\vx_a$. We can thus seek to repair this inherent structure by minimizing the loss associated with our multiple self-supervised task loss function $\mathcal{L}_m(\vx)$, defined as a weighted sum of three self-supervised task loss functions $\mathcal{L}_s(\vx), \mathcal{L}_i(\vx), \mathcal{L}_r(\vx)$, explained in detail in the following sections. To reverse an attack for a given input image $\vx'$ (which may or may not have been attacked), classifier $F$, norm parameter $p$ ($\infty$ here) and small $\epsilon_r$, we find reverse vector $\vr$, $$\vr = \arg\min_{\vr}  \mathcal{L}_m(\vx'+\vr) :  \|\vr\|_p<\epsilon$$

After finding a minimal $\vr$ via PGD we can recover robust classifications by classifying on $\vx'+\vr$ \citep{madry2018towards}. See \ref{algorithm: SSLattack} for the full reversal algorithm. Because $\mathcal{L}_m(\vx)$ is a sum of multiple objective functions, a whitebox attack will have to balance objectives, thus reducing its effectiveness as an attacker \citep{mao2020multitask}. As seen in the equation above, this approach to finding $\vr$ is independent of $F$, and is thus compatible with any classification architecture or supervised task.

\begin{figure}
    \centering
    \includegraphics[scale=0.35]{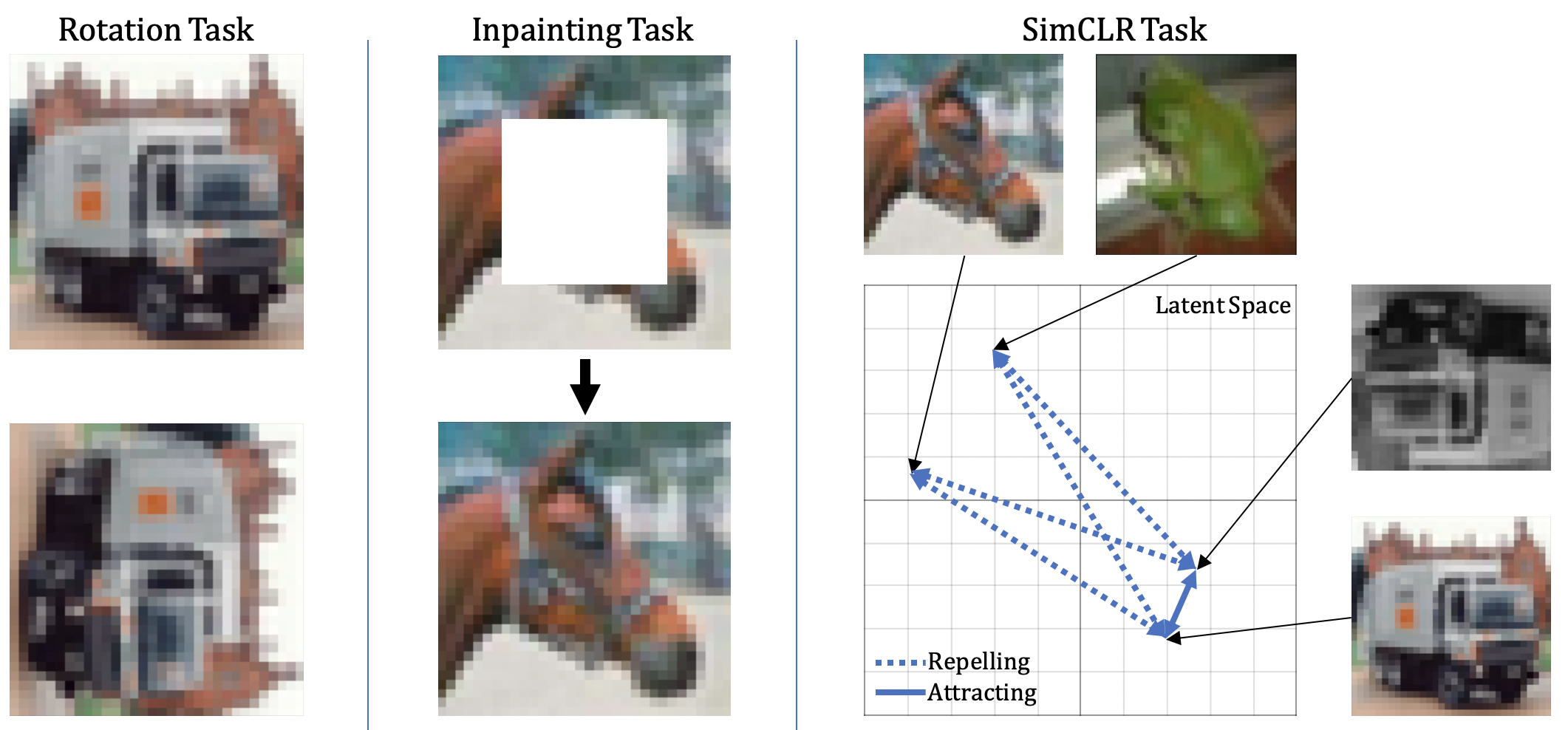}
    \caption{On the left see the rotation task, in which a network predicts the rotation of an image (lower left - 90 degree rotation, upper left - no rotation) \citep{gidaris2018unsupervised}. In the middle see the inpainting task in which given the upper middle image, a network attempts to predict the lower middle image \citep{pathak2016context}. On the right see the SimCLR task, in which a network tries to map augmented images (Crop, gray-scale and rotation shown here) from the same class closer to one another in a latent space than images from different classes \citep{chen2020simple}}
    \label{fig:my_label}
\end{figure}

\subsubsection{Contrastive Loss: $\mathcal{L}_s$}
In the SimCLR contrastive learning task introduced by \citet{chen2020simple}, we train a ResNet branch to minimize the latent space distance to copies of the same image under various transformations, and maximize the distance of non-matching image pairs, thus learning an augmentation-invariant representation of images \citep{he2015deep}. More formally, we define our contrastive loss for an image $\vx$ as $$ \mathcal{L}_s(\vx) = -\mathbb{E}_{i,j} \left[y_{i,j}\log\left(\frac{\exp(\vz_i\vz_j^T/\tau)}{\sum_k \exp(\vz_i\vz_k^T/\tau)}\right)\right]$$

Where $\vz_i$ is a possible result from transforming $\vx$,$\vz_j$ is a random transformed image, and $y_{i,j}$ is 1 if $\vz_i$ and $\vz_j$ originate from the same source image $\vx$, and 0 otherwise. $\tau$ is a hyperparameter. For our augmentations, we sequentially applied random cropping and color jittering, and also applied a horizontal flip and/or grayscale filter at random to each image. See \ref{fig:my_label} for an example.

\subsubsection{Inpainting Task Loss: $\mathcal{L}_i$}
In the inpainting task, we train an encoder-decoder deep network to fill in the missing centers of images, using encoders and decoder structures derived from the AlexNet architecture \citep{NIPS2012_c399862d, pathak2016context}. For a given whole image $\vx$, and context encoder-decoder $F$, we denote the output of $F$ on $\vx$ as $F(\vx)$. We define $\hat{M}$ to be a binary mask indicating dropped pixels via 1 and 0 otherwise. Letting $\odot$ define the element-wise product operation, we define the inpainting task loss on a given image $\vx$ as $$ \mathcal{L}_i(\vx) = \| \hat{M} \odot(\vx - F((1-\hat{M})\odot \vx)) \|_2^2$$ 

Intuitively speaking, we may interpret this as the $L_2$ norm of the difference between the true inpainted section of $\vx$, and the predicted inpainted section of the image returned by $F((1-\hat{M})\odot \vx)$ (predicted using all pixels but the pixels masked by $\hat{M}$, thus giving the $(1-\hat{M})$ term). See \ref{fig:my_label} for an example.

\subsubsection{Rotation Task Loss: $\mathcal{L}_r$}
In the image rotation task, we train a deep convolutional neural network to predict the rotation angle of an input image $\vx$. We use the experimental setup given in \citet{gidaris2018unsupervised}. Given a convolutional neural network $F$ with learnable parameters $\theta$ designed to predict image rotations, where $F^k(\vx)$ denotes the probability of $\vx$ having been transformed by rotation labeled $k$, and $\vx^k$ denotes $\vx$ transformed by rotation labeled $k$ we define the loss:$$\mathcal{L}_r(\vx,\theta) = - \frac{1}{K}\sum_{k=1}^K \log(F^k(\vx^k|\theta))$$
Intuitively, we treat this as a classification problem with cross-entropy loss and take the average loss across a set of K different rotations, trying to maximize the probability of the correct rotation for each $k$. Minimizing $\mathcal{L}_r(\vx,\theta)$ pushes $F^k(\vx^k|\theta) \rightarrow 1$ as desired. See \ref{fig:my_label} for an example.

\section{Results}

\begin{table}
\centering
\begin{tabular}{ |p{3cm}||p{2.1cm}|p{2.1cm}|  }
\hline
 \multicolumn{3}{|c|}{Classification Accuracy} \\
 \hline
 Reversal Method & Attacked Input& Clean Input\\
 \hline
 None   & 63.9\%    & 89.7\% \\
 SSL   & 65.3\%    & 86.6\%\\
 \textbf{MTL}&   \textbf{65.6\%}  & \textbf{87.7\%}\\
 \hline
\end{tabular}
\label{fig:results}
\caption{Using \citet{carmon2019unlabeled}'s robustly trained baseline model on CIFAR-10, we see a 1.1\% improvement in classification accuracy of clean images, and a 0.3\% improvement in classification accuracy of PGD attacked images compared with the state-of-the-art results from \citet{mao2021adversarial}}
\end{table}

In line with the literature, we set our attack bounds $\epsilon = 8$ \citep{madry2018towards}. In configuring reversal parameters  we continue \citet{mao2021adversarial}'s experiments and similarly find that the defense aware attacker can do no better than standard PGD attack, and that a reversal with $\epsilon = 8$ maximizes attacked input classification accuracy. Our results can are listed in \ref{fig:results} and our code can be found at \url{https://github.com/mattlawhon/SelfSupDefense/tree/one-by-one}.

\subsection{Analysis}
Using three self-supervised tasks, our approach yields significant improvement upon the state-of-the-art defense. On the CIFAR-10 dataset using \citet{carmon2019unlabeled}'s robustly trained baseline model, we achieved a 1.1\% improvement in classification accuracy of unattacked images, and a 0.3\% improvement in classification accuracy of PGD attacked images compared with the state-of-the-art results from \citet{mao2021adversarial}. Given an observed standard error of at most 0.4\% in the estimation of the approaches' true accuracy, this indicates statistically significant improvement in the classification accuracy of clean input with over 95\% confidence. Our work suggests that deep computer vision models' robustness can be improved by ensuring that they leverage the rich intrinsic structure of image data.

We note that this procedure increases computational cost because it requires the calculation of two gradients rather than one per iteration of the PGD reversal procedure. Further, we use different backbones for each self-supervised task, which improves accuracy of each task at the cost of performance. This may very well be an acceptable trade-off for better performance on clean input, though exploring this trade-off is an interesting direction for future research.

\section{Conclusions}

Though we show statistically significant improvements in overfitting over baseline, there is much additional work to be done in this domain because the scope of the experiments presented here is limited. Particularly exciting avenues for expanding this body of work include: experimenting with other self-supervised tasks, different datasets and adding a shared backbone for all self-supervised tasks for a true multi-task learning approach. Further, we note that this approach broadly furthers the finding that both multi-task learning and self-supervised learning hold promise to increasing adversarial robustness.

\subsection{Acknowledgements}
We would like to thank Gustave Ducrest for the productive discussions.

\bibliography{iclr2022_conference}
\bibliographystyle{iclr2022_conference}

\end{document}

%% file: math_commands.tex
\usepackage{amsmath,amsfonts,bm}

\def\eqref#1{equation~\ref{#1}}

\def\1{\bm{1}}

\def\vr{{\bm{r}}}

\def\vx{{\bm{x}}}

\def\vz{{\bm{z}}}

\DeclareMathAlphabet{\mathsfit}{\encodingdefault}{\sfdefault}{m}{sl}
\SetMathAlphabet{\mathsfit}{bold}{\encodingdefault}{\sfdefault}{bx}{n}

